%% file: main.tex
\documentclass{article}

\usepackage[preprint]{neurips_2025}

\usepackage[T1]{fontenc}    
\usepackage{hyperref}       
\usepackage{url}            
\usepackage{booktabs}       
\usepackage{amsfonts}       
\usepackage{amsmath}        
\usepackage{nicefrac}       
\usepackage{microtype}      
\usepackage{algorithm}
\usepackage{algpseudocode}
\usepackage{float}
\usepackage{enumitem}
\usepackage{booktabs}   
\usepackage{multirow}   
\usepackage{siunitx}    
\usepackage{colortbl}        
\usepackage[export]{adjustbox}
\usepackage{graphicx}
\usepackage{subcaption}
\usepackage{caption}
\usepackage{tabularx}

\newcolumntype{C}{>{\centering\arraybackslash}X}
\newcolumntype{L}{>{\raggedright\arraybackslash}X}

\usepackage{enumitem}

\usepackage{booktabs}       
\usepackage{amsfonts} 
\usepackage{amsmath}
\usepackage{wrapfig}
\usepackage{multirow}
\usepackage{colortbl}
\usepackage{multicol}
\usepackage{pifont}
\usepackage{makecell}  
\usepackage{xspace}

\usepackage{graphicx}
\usepackage{adjustbox}
\usepackage[table]{xcolor} 
\usepackage{amsmath}
\usepackage{amssymb}
\usepackage{lipsum} 
\usepackage{multirow}

\usepackage{siunitx}    
\usepackage{adjustbox}  
\usepackage{xfp}        

\definecolor{TopHeaderBlue}{HTML}{C5DFF8}       
\definecolor{SecondHeaderGray}{HTML}{E0E0E0}    
\definecolor{OurSectionHeaderBlue}{HTML}{A9CCE3} 
\definecolor{OurModelRowBlue}{HTML}{EBF5FB}     
\definecolor{DeltaRowPink}{HTML}{FADBD8}        

\definecolor{CloseSourceHeaderGreen}{HTML}{D5F5E3} 
\definecolor{OSGeneralHeaderYellow}{HTML}{FEF9E7} 
\definecolor{OSReasoningHeaderOrange}{HTML}{FDEBD0} 

\title{Empowering Lightweight MLLMs with Reasoning via Long CoT SFT}

\vspace{-10pt}

\author{%
  Linyu Ou \\
  Beijing Institute of Technology \\
  \texttt{zywoou@163.com} \\
  \And 
  Yuyang Yin \\
  Basic Algorithm Center, PCG, Tencent \\
  \texttt{yuyangyin@tencent.com} \\
}

\newcommand{\ignore}[1]{}

\begin{document}

\maketitle

\begin{center}
\vspace{-26pt}
  \textbf{Code Page:} \href{https://github.com/LinYuOu/mm-math-reason}{\textcolor{blue}{https://github.com/LinyuOu/mm-math-reason}}\\
  \textbf{Hugging Face:} \href{https://huggingface.co/TencentBAC/TBAC-VLR1-7B}{\textcolor{blue}{https://huggingface.co/TencentBAC/TBAC-VLR1-7B}}
\end{center}
\begin{abstract}
While Reinforcement Learning with Verifiable Rewards has enhanced the reasoning of large-scale language models (LLMs), its efficacy for lightweight multimodal language models (MLLMs) with fewer than seven billion parameters remains underexplored. This paper investigates the role of long Chain-of-Thought (long CoT) data in enhancing the reasoning abilities of such MLLMs. Our findings demonstrate that Supervised Fine-Tuning (SFT) with long CoT data significantly improves MLLM reasoning. Furthermore, we observe that after this initial SFT phase, MLLMs can achieve additional performance gains through a subsequent RL stage. We conclude that a SFT stage with long CoT data is a critical prerequisite for developing the reasoning capabilities of lightweight MLLMs.
\end{abstract}


\section{Introduction}
DeepSeek-R1 \cite{Deepseek-r1} has intensified researches on RL. Several studies \cite{VLAA-Thinker,Vlm-r1} adapt the standard SFT+RL pipeline to lightweight MLLMs. They use SFT as a cold-start stage and expect subsequent RL to substantially improve reasoning, yet Yue et al. \cite{RL-can't-beyond-the-base-model} demonstrate that RL does not create reasoning abilities but only incentivize them. The direct application of the SFT+RL pipeline to lightweight MLLMs yields only marginal reasoning gains\cite{MiMo-VL,Revisual-R1}, in contrast to the substantial gains observed in large-scale models.

Recent studies \cite{MiMo-VL,Revisual-R1} emphasize that long CoT SFT is critical for improving the reasoning capabilities of lightweight MLLMs. Our experiments similarly reveal that lightweight MLLMs struggle to engage in reflective reasoning. A case study, shown in Table~\ref{tab:grpo_vs_sft}, suggests that these models lack the foundational reasoning capacity required for RL to effectively amplify.

While most prior work focuses on models with 7B parameters or more, this study extends these findings by validating them on an even smaller 3B-parameter model and analyzing the impact of SFT data complexity on MLLM reasoning.
We further analyze the impact of SFT data complexity, demonstrating that reasoning gains are primarily driven by samples from the model's Zone of Proximal Development (ZPD)\cite{ZPD} that are challenging enough to provide a learning signal but still solvable.

In summary, our contributions are as follows:
\begin{itemize}[leftmargin=*, nosep]
    \item We validate the effectiveness of a long CoT SFT dataset for enhancing reasoning in both 7B and 3B parameters MLLMs, confirming its efficacy at lightweight models than previously studied.
    \item We are the first to systematically show that data difficulty, framed by the ZPD, is a critical factor for SFT efficiency, with challenging samples yielding the greatest performance improvements.
    \item We release our curated datasets and data generation code to facilitate future research.
\end{itemize}


\section{Related Work}

\subsection{Reinforcement Learning for Language Model Reasoning}

Aligning large language models (LLMs) with human preferences and enhancing their complex reasoning capabilities are central challenges in AI research. Foundational approaches like Reinforcement Learning from Human Feedback (RLHF), typically implemented via Proximal Policy Optimization (PPO), are powerful but suffer from high computational overhead and hyperparameter sensitivity. While alternatives like Direct Preference Optimization (DPO) simplify the process by using static preference data, their efficacy is limited in tasks requiring dynamic interaction.To address these limitations, recent work has produced specialized, reasoning-oriented RL algorithms. DeepSeek-R1, for instance, introduced Group Relative Policy Optimization (GRPO), which improves training stability and memory efficiency by using group-wise comparisons to estimate reward baselines. The subsequent Dr.GRPO variant further refines this approach by incorporating unbiased policy gradient calculations to mitigate length bias, thereby enhancing both token efficiency and reasoning accuracy \cite{Dr.GRPO}. Similarly, DAPO introduced a suite of techniques tailored for GRPO, including decoupled policy clipping and dynamic data sampling.
A parallel line of inquiry focuses on reward design within the RL framework. Methods like Test-Time Reinforcement Learning (TTRL) \cite{Ttrl} enable models to self-improve on unlabeled data by estimating rewards through majority voting. Complementing these algorithmic advances, other research highlights the profound impact of input formulation. For instance, a recent study demonstrated that simply adding the word "wait" to a prompt can trigger a model's latent reflective capabilities \cite{wait_prompt}, offering novel perspectives on LLM optimization. While these methods have shown success on large-scale models, their applicability to and efficacy on lightweight MLLMs with limited intrinsic reasoning abilities remain a significant, unexplored research area.

\subsection{Reinforcement Learning for Multimodal Reasoning}

The success of RL in text-based reasoning has inspired its application to the multimodal domain. Since the introduction of GRPO, a wave of multimodal studies has emerged-including VLM-R1 \cite{Vlm-r1}, Vision-R1 \cite{Vision-r1}, R1-OneVision \cite{R1-Onevision}, R1-Omni \cite{R1-Omni}, and Visual-RFT \cite{Visual-RFT} targeting a wide array of tasks from geometric reasoning to object detection. These works primarily focus on designing task-specific verifiable rewards, such as Intersection over Union (IoU) for spatial accuracy or classification correctness for image categorization, demonstrating the versatility of the RL framework.
Several studies specifically validate the effectiveness of GRPO in a multimodal context\cite{Lmm-r1, Revisual-R1}. For example, \cite{Lmm-r1} showed that applying GRPO with text-only data provided an initial performance boost, which was further enhanced by training on multimodal data. This "cold-start" strategy, where models are first warmed up on large volumes of text-based reasoning problems, has become a recurring theme.
While these findings highlight the importance of SFT, our work extends this line of inquiry in two critical directions. First, whereas prior studies have focused on bigger than 7B-scale models, we validate the efficacy of long CoT SFT on an even smaller 3B-parameter model. Second, we demonstrate that the difficulty of SFT data is a crucial factor for MLLM reasoning, establishing that the relatively challenging examples yield the greatest performance improvements. 

\section{Methodology}
\subsection{Dataset Construction}
\label{sec:dataset_construction}

To enhance the reasoning capabilities of lightweight MLLMs, we constructed a dataset of mathematical problems with fine-grained difficulty stratification. Our hypothesis is that training on a curated dataset of high-difficulty, long CoT problems is essential. This process serves two specific objectives: (1) preparing data with a difficulty range suitable for stable GRPO training, and (2) enabling a study of how data difficulty impacts model performance during SFT.

The data construction process began with a filtering stage, conceptually inspired by the educational theory of the ZPD \cite{ZPD}. The ZPD framework categorizes tasks relative to a learner's ability into three distinct zones: (1) tasks the learner can complete independently (the comfort zone), (2) tasks the learner can complete with guidance (the learning zone, or ZPD), and (3) tasks the learner cannot complete even with guidance (the frustration zone).

Our methodology operates on the assumption that a model's inference success rate over multiple attempts serves as a practical proxy for problem difficulty relative to its capabilities, thereby approximating its ZPD. To operationalize this, we used the baseline model qwen2.5-VL-7B to generate 16 inference attempts for each problem; this number was chosen as a trade-off between obtaining a stable estimate of the success rate and managing the computational cost of generation. We then excluded: (1) Problems where all 16 attempts were correct, analogous to the comfort zone (trivial tasks). (2) Problems where all 16 attempts were incorrect, analogous to the frustration zone (unsolvable tasks). 

The remaining problems, where the model's success rate is between 1/16 and 15/16, constitute our ZPD-approximated dataset. Our hypothesis is that by using this data, the training process can focus on problems that are challenging yet achievable, thus providing more informative learning signals for both SFT and subsequent GRPO training.
To establish a more granular difficulty metric within this ZPD, we used the number of correct outputs (from 1 to 15) as the definitive difficulty score for each problem. Based on these scores, the dataset was stratified into 15 discrete difficulty tiers.

This entire data construction protocol was applied to create the datasets for both training phases. The SFT dataset was primarily compiled from OpenR1-Math-220k \cite{openr1}. The dataset for the subsequent Reinforcement Learning (RL) phase was curated from various public sources and subjected to the identical filtering and stratification process. To ensure reproducibility, all constructed datasets have been made publicly available on Hugging Face.

\subsection{Training Pipeline} 
\label{sec:training_pipeline}

\subsubsection{Stage 1: Supervised Fine-Tuning for Enhancing Reasoning} 
The first stage of our pipeline involves SFT conducted exclusively on long CoT data. We adopted this text-centric approach based on the observation that the reasoning chains in many existing multimodal datasets are insufficient for cultivating robust reasoning in lightweight MLLMs. Specifically, we found their CoT sequences to be substantially shorter and less complex than those in high-quality, text-only datasets. As validated by our ablation studies (Section~\ref{sec:sft_ablation}), prioritizing long, text-based CoT proved to be a highly effective strategy.

\subsubsection{Stage 2: Reinforcement Learning with Modified GRPO}
In the second stage, we refined the SFT model using a modified Group Relative Policy Optimization (GRPO) strategy. Our key modifications to the standard framework include: single-step policy updates, removal of the KL divergence loss, our previously described difficulty-based data filtering, and clip-higher and overlong sequence filtering from DAPO \cite{Dapo}. We also adopted a simplified, accuracy-only reward structure, as the model had already learned format during the SFT stage, rendering format reward unnecessary.

The removal of the KL divergence loss is particularly noteworthy and is motivated by our data construction strategy. We hypothesize\footnote{This is a simplifying assumption whose validity requires further investigation.} that, by restricting training to problems within the model's ZPD, the risk of catastrophic policy shifts is inherently reduced. Under this assumption, the KL penalty-typically used as a stabilizing term-becomes less critical, enabling more aggressive policy optimization.

\section{Experiments}
\label{experiments}

\subsection{Experimental Setup}

\paragraph{Benchmarks} 
We validated our approach on several widely recognized mathematical reasoning benchmarks, including MathVision \cite{MathVision}, MathVerse \cite{MathVerse}, MathVista \cite{MathVista}, DynaMath \cite{DynaMath}, and OlympiadBench \cite{OlympiadBench}.

\paragraph{Evaluation}
We adopt a unified prompt instruction across all evaluations and require models to enclose their
final answers within ''\verb|\boxed{}|'', where the complete prompt is presented in Section \ref{prompt}. Model inference is conducted using vLLM\footnote{\url{https://github.com/vllm-project/vllm}} for accelerating generation. All questions are evaluated using Math-Verify\footnote{\url{https://github.com/huggingface/Math-Verify}}.

\paragraph{Implementation Details}
All experiments were conducted on NVIDIA H20 GPUs using the MS-Swift library\footnote{\url{https://github.com/modelscope/ms-swift}}. 
During SFT, we used a global batch size of 128 and set the maximum sequence length to 32,768 tokens to accommodate the full context of long-form reasoning problems.During the Reinforcement Learning (RL) phase, hyperparameters were adjusted for each model's scale. The batch sizes were configured to maximize training speed within the constraints of our available computational resources. While the SFT context window was 32,768 tokens, the maximum generation length during RL was set to 8,192. This was empirically determined to be sufficient for generating complete solutions while significantly reducing computational overhead during RL phase.
For Qwen2.5-VL-3B-Instruct, we generated 4 candidate responses per problem, used a global training batch size of 14, and trained for 300 steps.
For the larger Qwen2.5-VL-7B-Instruct, we trained for 40 steps with a global batch size of 48. We increased the number of candidates to 14 per problem to provide a more diverse set of responses for policy optimization, a strategy we found beneficial for the larger model. In contrast, 4 candidates were sufficient for stable training of the 3B model.

\subsection{Prompt Design}
\label{prompt}
The prompt used in training and inference:

\begin{quote}
\textbf{SYSTEM:} \\
You FIRST think about the reasoning process as an internal monologue step by step and then provide the final answer. \\
The reasoning process MUST BE enclosed within \texttt{<think></think>} tags. \\
The final answer MUST BE put in \verb|\boxed{}|.
\end{quote}

\subsection{Main Results}
The main result showed in Table\ref{tab:main_result} our model is named TBAC-VLR1, has 3B and 7B size.
The main results of our study are presented in Table \ref{tab:main_result}, which details a performance comparison of our TBAC-VLR1 model against other leading Vision-Language Models (VLMs) across a suite of mathematical and logical reasoning benchmarks. The results clearly demonstrate that our TBAC-VLR1 model achieves state-of-the-art (SOTA) overall performance in both the $\leq$ 3B and 7B parameter classes.

In the $\leq$ 3B model category, TBAC-VLR1-3B ranks first with an average score of 36.7, showcasing its superior comprehensive capabilities. The model achieves the highest scores on the challenging MathVerse (41.1) and MathVision (28.7) benchmarks and secures the second-best score on DynaMath (16.1). This highlights the significant effectiveness of our approach in enhancing the model's ability to tackle complex mathematical problems.

This leading performance extends to the 7B parameter class, where TBAC-VLR1-7B again secures the top position with an average score of 43.4. Notably, the model ranks first on three distinct benchmarks: MathVerse (50.1), MathVision (31.4), and DynaMath (22.6). While some models exhibit strong performance on a single benchmark (e.g., VL-Rethinker-7B on MathVista or VLAA-Thinker-Qwen2.5-7B on LogicVista), their performance is often compromised on other tasks. In contrast, TBAC-VLR1 demonstrates a more balanced and powerful profile. Its consistent high performance across diverse mathematical and logical tasks underscores its robustness and generalization capabilities as a universal math-reasoning model.

\input{table_main_result}
\subsection{Ablation Study}
\label{sec:ablation_studies}

This section details the ablation studies conducted to isolate the effects of different training components.

\subsubsection{Supervised Fine-Tuning Strategies}
\label{sec:sft_ablation}

We first evaluated the impact of data composition and difficulty-based sampling on model performance during SFT. The SFT datasets were curated from the \cite{openr1} and \cite{R1-Onevision}. All experiments were conducted using the Qwen2.5-VL-3B model on the MathVision benchmark, with a maximum response length of 32k tokens.

Our initial investigation focused on data modality and scale, with results detailed in Table~\ref{tab:sft_data_composition}. A model fine-tuned exclusively on 13,000 text-only instances with long-form Chain-of-Thought (CoT) reasoning achieved a baseline accuracy of 32.92\%. In contrast, a model trained on 12,000 multimodal instances yielded a significantly lower accuracy of 22.74\%. A mixed-modality dataset combining all 25,000 instances resulted in an accuracy of 32.81\%, showing no discernible improvement over the text-only configuration. These findings suggest that text-only CoT data provides a more effective training signal for this task. To validate this, we expanded the text-only dataset to 40,000 instances, which led to a notable performance gain, with the model achieving a final accuracy of 35.98\%.

Next, we analyzed the effect of training data difficulty. Leveraging the method described in Section~\ref{sec:dataset_construction}, we partitioned samples based on difficulty, quantified by a baseline model's inference performance. This yielded three 6,000-sample subsets: "Hard" (1--5 correct inferences), "Medium" (6--11), and "Easy" (12--15). As shown in Table~\ref{tab:sft_data_difficulty}, fine-tuning exclusively on the "Hard" subset produced the highest accuracy in this experiment (29.38\%), outperforming both the "Medium" (27.63\%) and "Easy" (27.83\%) subsets. Although this accuracy does not surpass that achieved with larger datasets, it demonstrates that prioritizing most challenging samples the model's ZPD is a highly data-efficient training strategy.

Although we observed superior performance with text-only data, this advantage may partly stem from the scarcity of high-quality multimodal reasoning datasets. Our core finding is that, regardless of modality, the length and complexity of the reasoning chain-i.e., data quality-are key to activating the reasoning capabilities of lightweight models. Future work could explore whether constructing multimodal datasets with equally high-quality reasoning chains can yield further improvements.

\subsubsection{Synergy of SFT and Reinforcement Learning}
\label{sec:sft_rl_synergy}

Finally, we investigated whether a long CoT SFT stage enhances the efficacy of subsequent Reinforcement Learning (RL). We compared the performance of models trained with RL alone against models that first underwent our long CoT SFT phase before RL. These experiments were conducted on both the Qwen2.5-VL 3B and 7B model variants.

The results, presented in Table~\ref{tab:rl_vs_sft_results_styled}, confirm that initializing the RL phase with a model already fine-tuned on long CoT data leads to significant performance gains in the final multimodal model. This demonstrates a clear synergistic relationship, where the foundational reasoning capabilities instilled by SFT are effectively amplified by the subsequent RL stage.

\input{table_w_o_sft}

\begin{table}[htbp]
  \centering
  \caption{Performance on the MathVision Benchmark by SFT Data Composition and Scaling.}
  \label{tab:sft_data_composition}
  \begin{tabular}{l c c}
    \toprule
    \textbf{SFT Data Configuration} & \textbf{Dataset Size} & \textbf{Accuracy (\%)} \\
    \midrule
    \multicolumn{3}{l}{\textit{Initial Composition Experiments}} \\
    Pure Text & 13,000 & 32.92 \\
    Multimodal & 12,000 & 22.74 \\
    Mixed (Text + Multimodal) & 25,000 & 32.81 \\
    \midrule
    \multicolumn{3}{l}{\textit{Text Data Scaling Experiments}} \\
    Pure Text (Baseline) & 13,000 & 32.92 \\
    Pure Text (Scaled) & 20,000 & 34.14 \\
    Pure Text (Scaled) & 40,000 & 35.98 \\
    \bottomrule
  \end{tabular}
\end{table}

\begin{table}[htbp]
  \centering
  \caption{Impact of SFT Data Difficulty on MathVision Benchmark Performance.}
  \label{tab:sft_data_difficulty}
  \begin{tabular}{l c c c}
    \toprule
    \textbf{Difficulty Bin} & \textbf{Definition (Correct Inferences)} & \textbf{Subset Size} & \textbf{Accuracy (\%)} \\
    \midrule
    Hard & 1--5 & 6,000 & 29.38 \\
    Medium & 6--11 & 6,000 & 27.63 \\
    Easy & 12--15 & 6,000 & 27.83 \\
    \bottomrule
  \end{tabular}
\end{table}

\subsection{Case Study}
Table~\ref{tab:grpo_vs_sft} demonstrates that, after SFT on long chain-of-thought (CoT) data, the \texttt{Qwen2.5-VL-3B} model is more likely to generate tokens such as ''wait,'' indicating a tendency to pause, reconsider, and refine its reasoning process. This suggests that lightweight multimodal large language models (MLLMs) can readily acquire reasoning and reflection capabilities through SFT. In contrast, RL exhibits limited effectiveness in cultivating these abilities.

\begin{table}[htbp]
  \centering
  \begin{tabular}{@{}p{0.2\linewidth}p{0.75\linewidth}@{}}
    \toprule
    \multicolumn{2}{@{}p{0.95\linewidth}@{}}{
      \centering
      \includegraphics[width=0.5\linewidth]{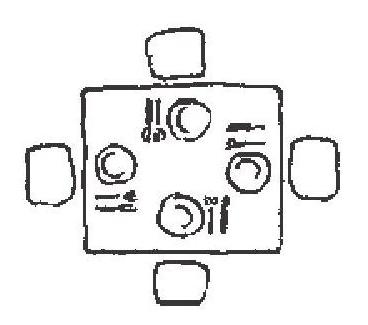}\\[0.5em]
      \textbf{Problem:} Four people can be seated at a square table. How many people at most could be seated if we pushed four tables of this kind together in one row?\\[0.3em]
      \textbf{Answer:} 10
    } \\[0.5em]
    \midrule
    \textbf{Model} & \textbf{Response} \\
    \midrule
    \textbf{After GRPO} &
    \begin{minipage}[t]{0.75\linewidth}
      \vspace{2pt}
      \small
      To solve this problem, I need to visualize how the tables can be pushed together.
      
      If we push four square tables together in one row, we can imagine them as a single large rectangle. Each table has four sides, but when pushed together, the sides that touch each other will not be available for seating.
      
      So, each table contributes three sides for seating. Since there are four tables, we multiply the number of tables by the number of sides each contributes to seating:
      
      \centerline{$4 \text{ tables} \times 3 \text{ sides per table} = 12 \text{ sides}$}
      
      Therefore, the maximum number of people that could be seated is $\boxed{12}$.
      \vspace{2pt}
    \end{minipage} \\[1em]
    
    \textbf{After SFT} &
    \begin{minipage}[t]{0.75\linewidth}
      \vspace{2pt}
      \small
      Okay, I need to figure out how many people can be seated if we push four of these square tables together in a row.
      Each table can seat four people, right? \textcolor{red}{\textbf{Wait}} \ldots If they're \ldots \textcolor{red}{\textbf{Wait}} \ldots the total seats would be $(4 + 4 - 2) = 6$ seats? \textcolor{red}{\textbf{Wait}}, maybe not. Let's see: each table has four seats \ldots
      
      \textcolor{red}{\textbf{Wait}} \ldots if they are in a straight line, each internal edge is shared between two tables, but the ends are open. \textcolor{red}{\textbf{Wait}} \ldots perhaps there's another arrangement so that more seats are available?
      
      For example, \ldots When pushed together, the shared edge is a side of the square, which is a single seat? \textcolor{red}{\textbf{Wait}} \ldots maybe each table has four seats, each at the midpoint of each side? No, that would make eight seats. \textcolor{red}{\textbf{Wait}} \ldots the total number of people that could be seated is $\boxed{10}$.
      \vspace{2pt}
    \end{minipage} \\
    \bottomrule
  \end{tabular}
  \caption{Qualitative comparison between GRPO and SFT approaches for the table seating problem.}
  \label{tab:grpo_vs_sft}
\end{table}

\section{Conclusion}
\label{sec:conclusion}

This work challenges the prevailing assumption that SFT+RL pipelines scale effectively to lightweight MLLMs. We find that for these models, simply scaling down the pipeline yields only marginal gains, necessitating a paradigm shift that prioritizes SFT.
We demonstrate that an intensive SFT phase using high-quality, long CoT data is the foundational prerequisite for instilling robust reasoning. Our key insights are threefold: (1) Long CoT data is the primary driver of reasoning ability; (2) Data difficulty, framed by the ZPD, is critical for learning efficiency; and (3) SFT builds the core capability that subsequent RL can then effectively refine, confirming a strong synergy.
Our work establishes a principle for lightweight models: data quality is paramount. We conclude that future efforts should prioritize high-quality SFT data curation as the primary mechanism for capability development, with RL serving as a powerful but secondary optimization tool. To catalyze further research, we release our models, curated datasets, and source code to the community.

\clearpage

\bibliography{references}
\bibliographystyle{unsrt}


\end{document}

%% file: table_main_result.tex
\begin{table}[!htbp]
  \small
  \setlength{\tabcolsep}{3pt}
  \renewcommand{\arraystretch}{1.4} 

  \caption{Performance comparison of various models on math-related benchmarks.
           The best scores are \textbf{bold}; the second best are \underline{underlined} (among listed models).
           Highlighting is performed separately for models $\le$3B and 7B.
  }
  \label{tab:main_result}

  \begin{adjustbox}{width=\textwidth,center}
  \begin{tabular}{lcccccc}
    \toprule
    \rowcolor{TopHeaderBlue}
    & \multicolumn{6}{c}{\textbf{Math and Logic Benchmarks}} \\
    \cmidrule(lr){2-7}
    \rowcolor{SecondHeaderGray}
    \textbf{Model} & \textbf{MathVerse} & \textbf{MathVision} & \textbf{MathVista} & \textbf{DynaMath} & \textbf{LogicVista} & \textbf{Avg.} \\
    \midrule
    \multicolumn{7}{l}{\textit{Models $\le$ 3B}} \\
    \midrule
    Qwen2-VL-2B                     & 17.5 & 16.1 & 48.0 &  3.8 & 26.6 & 22.4 \\ 
    InternVL2.5-2B                  & 22.3 & 14.0 & 51.1 &  4.4 & 27.3 & 23.8 \\ 
    InternVL3-2B                    & 24.5 & 20.2 & 57.6 & 14.8 & \underline{40.3} & 31.5 \\ 
    Qwen2.5-VL-3B                   & 31.2 & 21.9 & 61.2 & 13.2 & \underline{40.3} & 33.6 \\ 
    VLM-R1-3B-Math-0305             & 32.2 & 21.9 & \underline{62.7} & 13.0 & \textbf{40.5} & 34.1 \\ 
    Taichu-VLR-3B                   & 32.1 & 23.1 & \textbf{64.9} & 12.6 & 38.7 & 34.3 \\ 
    VLAA-Thinker-Qwen2.5VL-3B       & \underline{36.4} & \underline{24.4} & 61.0 & \textbf{18.2} & 38.5 & \underline{35.7} \\ 
    TBAC-VLR1-3B                    & \textbf{41.1} & \textbf{28.7} & 57.5 & \underline{16.1} & 40.0 & \textbf{36.7} \\ 
    \midrule
    \multicolumn{7}{l}{\textit{Models = 7B}} \\
    \midrule
    Qwen2.5-VL-7B                       & 45.5 & 25.7 & \underline{68.0} & 21.8 & 41.2 & 40.5 \\ 
    VLAA-Thinker-Qwen2.5-7B             & \underline{48.2} & 26.4 & \underline{68.0} & \underline{22.4} & \textbf{48.5} & \underline{42.7} \\
    VL-Rethinker-7B                     & 46.4 & \underline{28.4} & \textbf{73.7} & 17.8 & 42.7 & 41.8 \\
    TBAC-VLR1-7B                        & \textbf{50.1} & \textbf{31.4} & 66.7 & \textbf{22.6} & \underline{46.4} & \textbf{43.4} \\ 
    \bottomrule
  \end{tabular}
  \end{adjustbox}
\end{table}

%% file: table_w_o_sft.tex
\begin{table}[!htbp]
  \small 
  \setlength{\tabcolsep}{3pt} 
  \renewcommand{\arraystretch}{1.4} 

  \caption{
      Performance comparison of Reinforcement Learning (RL) with and without Supervised Fine-Tuning (SFT). 
      For each model group, the highest score is shown in \textbf{bold}.
  }
  \label{tab:rl_vs_sft_results_styled} 

  \begin{adjustbox}{width=\textwidth,center}
  \begin{tabular}{lcccccc}
    \toprule
    & \multicolumn{6}{c}{\textbf{Benchmarks}}\\
    \cmidrule(lr){2-7} 

    \textbf{Model} & \textbf{MathVerse} & \textbf{MathVision} & \textbf{MathVista} & \textbf{DynaMath} & \textbf{LogicVista} & \textbf{Average} \\
    \midrule

    \multicolumn{7}{l}{\texttt{3B}} \\
    \midrule
    RL w/o SFT & 40.4          & 25.8          & \textbf{58.2} & \textbf{17.7} & 38.7          & 36.1 \\ 
    RL w/ SFT  & \textbf{41.1} & \textbf{28.7} & 57.5          & 16.1          & \textbf{40.0} & \textbf{36.7} \\ 
    \midrule

    \multicolumn{7}{l}{\texttt{7B}} \\
    \midrule
    RL w/o SFT & 43.4          & 25.4          & \textbf{70.1} & 19.0          & \textbf{48.4} & 41.3 \\ 
    RL w/ SFT  & \textbf{50.1} & \textbf{31.4} & 66.7          & \textbf{22.6} & 46.4          & \textbf{43.4} \\
    \bottomrule
  \end{tabular}
  \end{adjustbox}
\end{table}